\documentclass[runningheads,12pt]{llncs}
\usepackage[margin=3cm]{geometry}
\usepackage{amsmath,amsfonts,amssymb}
\usepackage{graphicx}
\usepackage[strings]{underscore}
\usepackage{hyperref}
\hypersetup{
	breaklinks=true,
	colorlinks=true,
	urlcolor=purple,
	linkcolor=blue,
	citecolor=blue
	}
\newcommand{\mg}[1]{\mathbf{#1}}
\usepackage{booktabs}

\usepackage{tabu}

\usepackage{longtable}
\usepackage{rotating,tabularx}
\usepackage{multicol}
\usepackage{multirow}
\usepackage{threeparttablex}

\usepackage{xcolor}
\definecolor{somegray}{rgb}{0.5, 0.5, 0.5}
\newcommand{\darkgrayed}[1]{\textcolor{somegray}{#1}}

\begin{document}

\title{Visual Radial Basis Q-Network}

\author{Julien Hautot\inst{1} \and
Céline Teuliere\inst{1} \and
Nourddine Azzaoui\inst{2}}
\institute{Université Clermont Auvergne, Clermont Auvergne INP, CNRS, Institut Pascal, F-63000 Clermont-Ferrand, France 
\email{\{julien.hautot,celine.teuliere\}@uca.fr} \and
Laboratoire de Mathématiques Blaise Pascal, Clermont-Ferrand 63100, France\\
\email{azzaoui.nourddine@uca.fr}}
\maketitle              %
\begin{abstract}

While reinforcement learning (RL) from raw images has been largely investigated in the last decade, existing approaches still suffer from a number of constraints. The high input dimension is often handled using either expert knowledge to extract handcrafted features or environment encoding through convolutional networks. Both solutions require numerous parameters to be optimized. In contrast, we propose a generic method to extract sparse features from raw images with few trainable parameters. We achieved this using a Radial Basis Function Network (RBFN) directly on raw image. We evaluate the performance of the proposed approach for visual extraction in Q-learning tasks in the Vizdoom environment. Then, we compare our results with two Deep Q-Network, one trained directly on images and another one trained on feature extracted by a pretrained auto-encoder. We show that the proposed approach provides similar or, in some cases, even better performances with fewer trainable parameters while being conceptually simpler.

\keywords{Reinforcement Learning  \and State Representation \and Radial Basis Function Network \and Computer Vision}
\end{abstract}
\textit{\small\darkgrayed{This paper has been accepted for publication at the	
		3rd International Conference on Pattern Recognition and Artificial Intelligence, ICPRAI 2022. \copyright Springer Nature\\ DOI: \url{https://doi.org/10.1007/978-3-031-09282-4_27}}
		}
\begin{figure*}[t]
    \label{archi}
	\includegraphics[width=\textwidth,height=4cm]{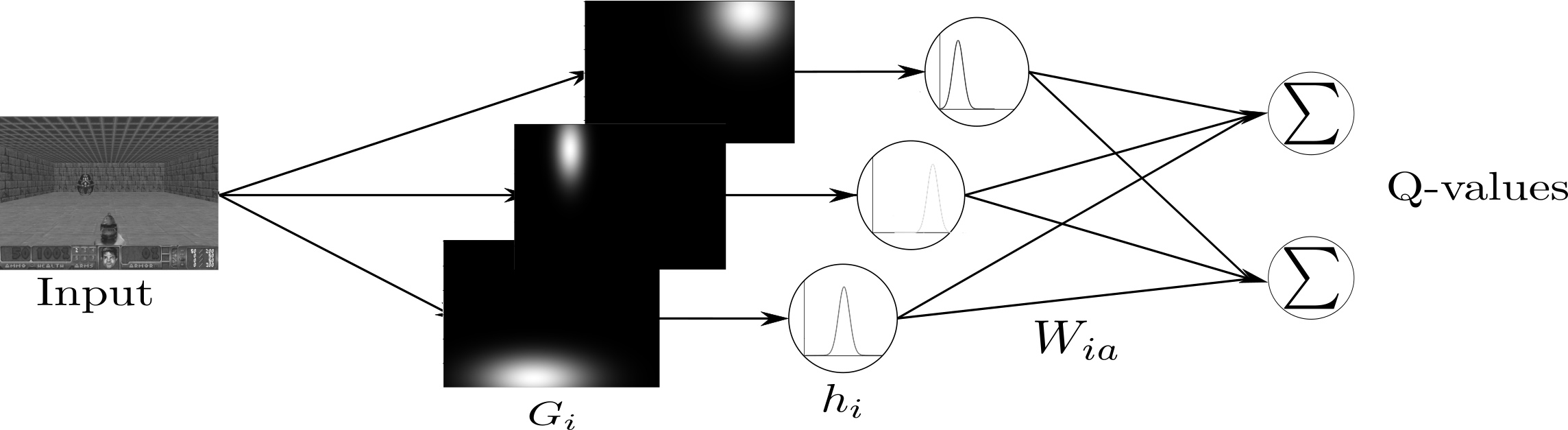}
	\caption{Visual Radial Basis Q-Network architecture. Input raw image are filtered by spatial Gaussian filters $G_i$ then Gaussian activation $h_i$ are applied on all the filtered pixels and Q-values are calculated by a linear combination of $h_i$ and $W_ia$ the weight between activation $h_i$ and output $Q(S,a)$}
\end{figure*}

\section{Introduction}
Reinforcement learning (RL) has made significant progress in recent years, allowing the use of policy and value-based algorithms on images to perform tasks in fully observable contexts~\cite{Christodoulou2019,Mnih2015,Schulman2017}. However, when the state space is more complex or partially observable, learning remains harder. Solutions are introduced by implementing additional learning modules like Long Short Term Memory (LSTM)~\cite{Hochreiter1997} or enlarging the networks by adding convolutional layers~\cite{Justesen2020}.\\
Another strategy to facilitate the learning of an agent in a complex visual environment is to reduce the dimension of the state space~\cite{Lesort2018}. This can be accomplished by extracting relevant features from the images using learning based methods such as auto-encoder~\cite{Lange2010}. The dimension of the state space has a large impact on how these solutions are implemented. Indeed, an auto-encoder needs lots of parameters to encode large states and hence its training is time and energy consuming.\\
Furthermore using sparse representation in RL has been proven to be efficient~\cite{Liu2019}, this can be done with Radial Basis Function Networks (RBFN)~\cite{BROOMHEAD1988,Hartman1990,Moody1988,Park1993} a well know function approximator used for classification and regression tasks~\cite{Wu2012}. Since its success in deep learning, the idea of using RBFN in RL has emerged and encountered a great interest especially in non-visual tasks applications~\cite{Asadi2021,Capel2020,Cetina2008}. In all these works, RBFN are however not applied directly on raw pixel, but rather on features extracted from the environment that involve pre-training steps.

\paragraph{Contributions.} In this paper, we aim to train an agent in a partially observable environment, with only raw pixel inputs and without prior knowledge, i.e., without pre-training or additional information. We extend the RBFN to extract random sparse features from visual images. Our major contribution is the design and the analysis of a generic features extraction method based on RBFN for visual inputs. We evaluate our extracted features performance in a Q-learning setting and compare the result with state-of-the-art methods. We used Vizdoom as the virtual environment~\cite{Kempka2016} where visual tasks are hard to solve with classic RL algorithms~\cite{Khan2020}.

\paragraph{Paper Organisation.} In section \ref{sec1}, we present a brief review of related approaches, then we introduce the theoretical background in section \ref{sec2}. In section \ref{sec3} we present our RBFN method and analyse the extracted features in section \ref{sec4}. Finally section \ref{sec5} presents an evaluation of the method on two different visual RL tasks and compare the performances against state-of-the-art approaches.

\section{Related Works}
\label{sec1}

State representation learning methods are popular approaches to reduce the state dimension used in visual RL. In the survey~\cite{Lesort2018}, Lesort et al. distinguished several categories of approaches for state representation learning. Auto-encoders can be used to extract features by reconstructing input images~\cite{Lange2010,Nair2018,Van2016}, or by forward model learning to predict the next state given the actual state and the corresponding action~\cite{Leibfried2018}. Learning an inverse model to predict the action knowing states can also be used to extract features that are unaffected by uncontrollable aspect of the environment~\cite{Zhong2020}. Prior knowledge can also be used to constrain the state space and extract conditional features~\cite{Finn2016}.
\noindent These learning representation methods typically require to train a network upstream of an RL algorithm, which can be time-consuming due to the large number of parameters or the environment dimension. Even when the state representation is trained parallel with the agent as in~\cite{Pathak2017} or when unsupervised learning is used as in~\cite{Anand2019,Srinivas2020}, the training is still time-consuming since multiple convolutional networks are trained. Unlike the current state-of-the-art, our method does not rely on pre-training or convolutional network; instead the features are extracted using a combination of Radial Basis Function (RBF) without training nor prior knowledge.\\
RBFNs have been recently used as feature extractors for 3-D point cloud object recognition~\cite{Chen2018}. They achieved similar or even better performances compared to the state-of-the-art with a faster training speed. Earlier, in~\cite{Daoud2019} RBF-based auto-encoder showed the efficiency of RBFN as feature extractors prior to classification tasks.
One key property of RBFN activation is their sparsity; for each input, only a few neurons will be activated. This is advantageous for RL as it imposes locality of activation avoiding catastrophic interference and keeping stable value which allows bootstrapping~\cite{Liu2019}. Applying directly RBFN on high dimensional images is challenging and has not been yet much explored in the literature. Other sparse representations, such as extreme machine learning or sparse coding, have been investigated but none have been extended to visual RL as their computing cost tends grows exponentially with the input dimension. One solution in the literature is to reduce the inputs dimension with convolution~\cite{Liu2018,Rodrigues2021}.

\section{Background}
\label{sec2}
\subsection{Reinforcement Learning}

In RL we consider an agent interacting with an environment through actions $a$, states $s$ and rewards $r$. The environment is modelled as a Markov decision process (MDP). Such processes are defined by the tuple $ \left\langle{S,A,T,R}\right\rangle $, where $S$ and $A$ are the states and action spaces, $T : S \times A \rightarrow S$ is the transition probability to go from a state $s\in S$ to another state $s'\in S$ by performing an action $a\in A$. $R : S \times A \rightarrow \mathbb{R}$ is the reward function. At each time step, the agent will face a state $s\in S$ and choose an action $a \in A$ according to its policy $\pi : S \rightarrow A$. The environment will give back a reward $r=R(s,a)$ and the probability of reaching $s'$ as the next state is given by $T(s,a,s')$. The goal of the agent is to find a policy which maximizes its expected return, which we define as the cumulative discounted reward along time-step $t$, i.e., $R_t = \sum_{k=0}^{\infty}\gamma^k r_{t+k+1} $, where $\gamma$ is the discounted rate determining the actual value of future reward. \\
We chose to evaluate our state representation method in a Q-learning~\cite{Sutton2018} setting. To optimize the policy, Q-learning estimates the state-action value function (Q-value) which represents the quality of taking an action, $a \in A$, in a state, $s \in S$, following a policy $\pi$. $Q^{\pi}(s,a)$ is defined by

\begin{equation}
Q^{\pi}(s,a) := \mathbb{E}_{\pi}[R_t | s_t=s, a_t=a].
\end{equation}

The optimal policy can then be estimated by approximating the optimal Q-value given by the Bellman equation such that

\begin{equation}
Q^*(s,a) = \mathbb{E}_{s'}[r + \gamma \max_{a'}Q^*(s',a')].
\label{Bellman}
\end{equation}

In Deep Q-learning~\cite{Mnih2015}, the optimal Q-value is approximated by a neural network parametrized by $\theta$.
At each time-step Q-values are predicted by the network, then the agent chooses the best action according to an exploration strategy. The agent experiences $\left\langle s_t,a_t,s_{t+1},F \right\rangle$ are stored into a replay memory buffer $D$ where $F$ is a Boolean indicating if the state is final or not. Parameters of the Q-Network $\theta$ are optimized at each iteration $i$ to minimize the loss $L_i(\theta_i)$ for a batch of state uniformly chosen in the replay buffer $D$, defined by
\begin{equation}
    L_i(\theta_i) = \left\{
    \begin{array}{ll}
    \mathbb{E}_{(s,a,r,s')\sim U(D)}[\left(r -  Q(s,a;\theta_{i})\right)^2] \text{ \: if \: F}\\
    \mathbb{E}_{(s,a,r,s')\sim Rb}[\left(r+\gamma\max_{a'}Q(s',a';\bar{\theta_i})-Q(s,a;\theta_{i})\right)^2]\text{ \: otherwise.}
    \end{array}\right.
    \label{loss}
\end{equation}

To stabilize the training a target network is used with parameters $\bar{\theta}$ which is periodically copied from the reference network with parameters $\theta$.

\subsection{Radial Basis Function Network (RBFN)} 

The network is composed of three layers: the inputs $\mg{X}$, the hidden layers $\phi$ compose of RBFs and the output layers $\mg{Y}$.
A RBF is a function $\phi$ defined by its center $\mu$ and its width $\sigma$ as
\begin{equation}
    \phi(\bf X) = \phi(||X- \mg{\mu})||, \mg{\sigma}), \text{\quad where \:} ||.|| \text{\: is a norm.}
\end{equation}

\noindent In this paper we will consider Gaussian RBF with the Euclidean norm defined as

\begin{equation}
\label{Gaussian}
\phi_g(\mg{X})=\exp\left(\frac{||\mg{X}- \mg{\mu}||^2}{2\mg{\sigma}^2}\right).
\end{equation}

\noindent A RBFN computes a linear combination of $N$ RBF:

\begin{equation}
    y_i(\mg{X}) = \sum_{j=1}^{N}w_{ij}\cdot \phi(||\boldsymbol{X}-\mg{\mu}_j||,\mg{\sigma}_j),
\end{equation}

\noindent where, $w_{ij}$ is a learnable weight between output $y_i$ and hidden RBF layer $\phi_j$.\\

RBFN are fast to train due to the well-localized receptive fields that allow activation of few neurons for one input. The farther the input value is from the receptive field of the neuron, the closer output value is to zero.

\section{Method}
\label{sec3}
Our method focuses on the projection of high dimensional input state spaces into a lower dimensional space by combining Gaussian receptive filters and RBF Gaussian activations. Our network architecture is shown in Fig \ref{archi}.
Each receptive field can be seen as the attention area of the corresponding neuron. The attention area of a neuron $i$, for a pixel $p$ in a state $S$ (of shape $w\times h$) with coordinates $(p_x,p_y)$ is defined as follows:
\begin{equation}
\label{receptivefield}
\begin{aligned}
G_{i,p_x,p_y} = \exp\left(-\left(\frac{(p_x/w- \mu_{x,i})^2}{2\sigma_{x,i}^2} +\frac{(p_y/h - \mu_{y,i})^2}{2\sigma_{y,i}^2}\right)\right),\\
\end{aligned}
\end{equation}

\noindent where, $\mu_{x,i},\mu_{y,i} \in [0,1]$ define the center of the Gaussian function along spatial dimension and $\sigma_{x,i}, \sigma_{y,i} \in [0,1]$ are the standard deviations. $\mg{G_i} \in \mathcal{M}_{w \times h}$ is the full matrix that defines the spatial attention of a neuron.
Given the attention area, the activation of the hidden neuron $i$ is computed using a Gaussian RBF activation function weighted by $\mg{G_i}$:

\begin{equation}
\label{Gaussianactivation}
\begin{aligned}
\mg h_i({\mg{S}}) = \exp\left(- \frac{\sum((\mg S-\mu_{z,i})\odot\mg G_{i})^2}{2\sigma_{z,i}^2}\right), \\
\end{aligned}
\end{equation}

\noindent where, $\mu_{z,i} \in [0,1]$ is the center and $\sigma_{z,i} \in [0,1]$ the standard deviation of the RBF intensity Gaussian activation function. Symbol $\odot$ is the Hadamard product, i.e., the element-wise product. Parameters $\mu_{z,i}$ and $\sigma_{z,i}$ have the same size as the input channel.

To test the efficiency of our extraction we use the extracted features as the state in a Q-learning algorithms where the Q-value will be approximated by a linear combination of $N_g$ Gaussian neurons activations.

\begin{equation}
    Q(s,a) = \sum_{i=0}^{N_g} w_{ai} \times h_i(s),
\end{equation}

\noindent Where, $w_{ai}$ is the weight between the action $a$ and the neuron $i$. On each step the input image is passed to the RBF layer and the computed features are saved in the replay buffer. Then during each training iteration a batch of random features is chosen from the replay memory buffer and the weights are updated using a gradient descent step (Adam~\cite{Kingma2015}) to minimize equation (\ref{loss}).

\subsection{Selection of RBF Parameters}

There are 6 hyper-parameters for each Gaussian unit. Centers of Gaussian filters $\mu_{x,y}$ and centers of Gaussian activation $\mu_z$ are chosen uniformly between $0$ and $1$ as we need to cover all the state space. The standard deviations $\sigma_{x,y,z}$ influence the precision of the activation of a neuron, i.e., the proximity between pixel intensities weighted by attention area and intensity center of the neuron. In that way RBF layer allows activation of few neurons for a particular image.\\ The hyper-parameters are chosen at the beginning and never changed during training.\\
After empirical experiments and based on the work of~\cite{Buessler2014} which used also Gaussian attention area, for all the study we choose $2001$ neurons for gray and $667$ neurons for rgb inputs as each neuron has 3 intensity center, one for each canal. The choice of standard deviations is as follows: $\sigma_z =1$ and $\sigma_{x,y}$ uniformly chosen in $[0.02-0.2]$.

\subsection{Vizdoom Scenarios}

We evaluate our method on two Vizdoom scenarios to show the robustness of our network to different state dimensions and partially observable tasks. Scenarios are defined as follows:

\subsubsection{Basic Scenario.}
In this task the agent is in a square room, its goal is to shoot a monster placed on a line in front of him. At each episode the monster has a different position while the agent spawns all the time in the same place. In this scenario the agent has the choice between 8 possible actions: move right, move left, shoot, turn left, turn right, move forward, move backward and do nothing. In this configuration the agent can be in a state where the monster is not visible. The agent gets a reward of 101 when the shoot hits the monster, -5 when the monster is missed and -1 for each iteration. The episode is finished when the agent shoots the monster or when it reaches the timeout of 300 steps.

\subsubsection{Health Gathering Scenario.}
In this task the agent is still in a square room but the goal is different, the agent has to collect health packs to survive. It has $100$ life points at the beginning of the episode, each iteration the agent looses some life points, if the agent reaches a health pack it gains some life point. The episode ends when the agent dies or when it reaches 2100 steps. Reward is designed as follows: $r_t = \mathrm{life}_{t+1} - \mathrm{life}_{t} $. Possible actions are move forward, move backward, turn left, turn right, do nothing. All the health packs spawn randomly and the agent spawns in the middle of the room. 

\section{Analysis of Pattern Activations}
\label{sec4}
In this section we put in evidence the sparsity of our network, and analyse the activation patterns of our neurons and their differences. We then present an example of feature targeted by a RBF neuron.

\begin{table}[!t]
	\caption{
		Distribution of the active neurons on 1000 states with 20 different RBF parameters initialisation. Per cent of the total number of neurons (2001).	 
	}
	\label{tab:dstrib}
	\begin{center}%
	\begin{tabu}{l cc cc}
		\toprule
		\multirow{2}{*}{} & \multicolumn{2}{c}{\texttt{Basic}} & \multicolumn{2}{c}{\texttt{Health gathering}}\\
		\cline{2-5}
		& \texttt{gray} & \texttt{RGB} & \texttt{gray} & \texttt{RGB}\\
		\midrule
		\midrule 
		\texttt{active neurons (aN)} & $32 \pm 1 \%$ & $31 \pm 1 \%$ & $29 \pm 6 \%$ & $32 \pm 5 \%$ \\
		\bottomrule
	\end{tabu}
\end{center}
\end{table}

\begin{figure}[h]
	\centering
	\includegraphics[width=\textwidth,height=6cm]{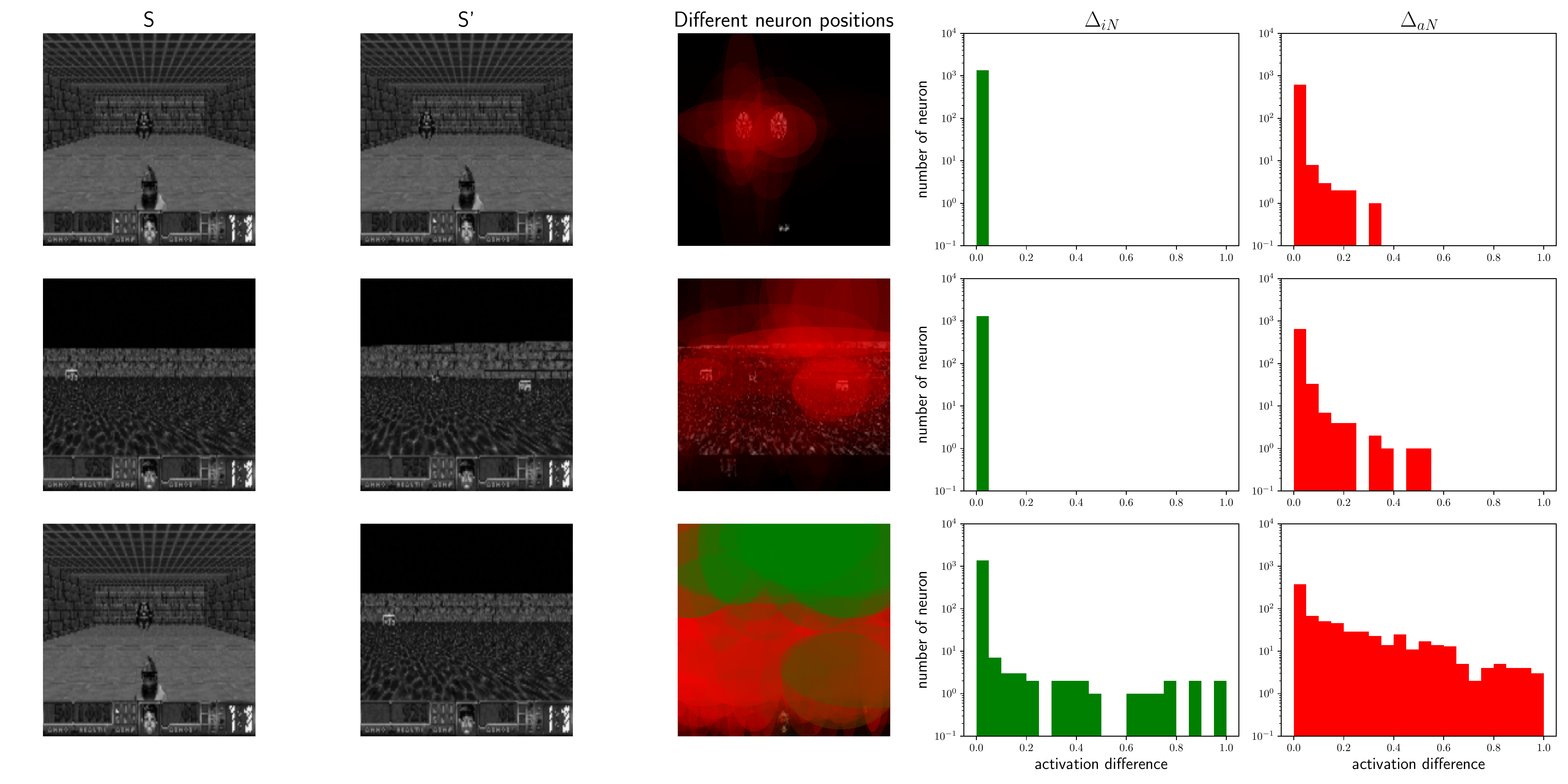}
	\caption{Comparison of neurons activity on different states for one seed. Histograms represent $\Delta_N = |N(s)-N(s')|$ with N the neurons activation (iN,aN). Neuron positions represent the position of the different neurons with $|S-S'|$ in the background. \textit{Top:} comparison for two close states in basic scenario. \textit{Middle:} comparison of two close states in the health gathering scenario. \textit{Bottom:} comparison between both scenarios}
	\label{stm}
\end{figure}

\noindent We generated 1 000 states from both scenarios in gray and rgb with a random policy and a random "skip frame" between 0 and 12 in order to cover all the state space, i.e., each random action is repeated a certain number of time. In each scenario we can classify the neurons into two categories: inactive neurons (iN) which have an activation always inferior to $0.01$ and active neurons (aN) which have an activation superior to $0.01$ in some states. Table \ref{tab:dstrib} gives the percentage of aN on the total number of neurons for 20 different seeds in each configuration. We can see that even with a random generation of Gaussian neurons we have stable distribution, around $30\%$ of activated neurons, across different random seeds, state spaces and scenarios.

Fig \ref{stm} highlights the difference in activation for iN and aN between two frames. The first observation is that iN can be used to differentiate two scenarios as some iN of basic scenario become aN in health gathering scenario, those neurons describe the scenario rather than the state. The active neurons, have also more differences between two scenarios than two states, but they have a non negligible number of neurons that differ between the states. These neurons describe particular features as shown in the neuron position column. Indeed, in the basic scenario, the position of different aNs are around the monster, for health gathering there is more differences between both frames so more different neurons. However, we identified 3 main areas, two on both health packs and one on the upper side of the right wall. Between both scenarios, the different iNs and aNs are distributed over the entire image with more differences on the top for the iN as the ceiling is characteristic of the scenario.

\begin{figure}[h!]
	\centering
    \includegraphics[width=\textwidth,height=6cm]{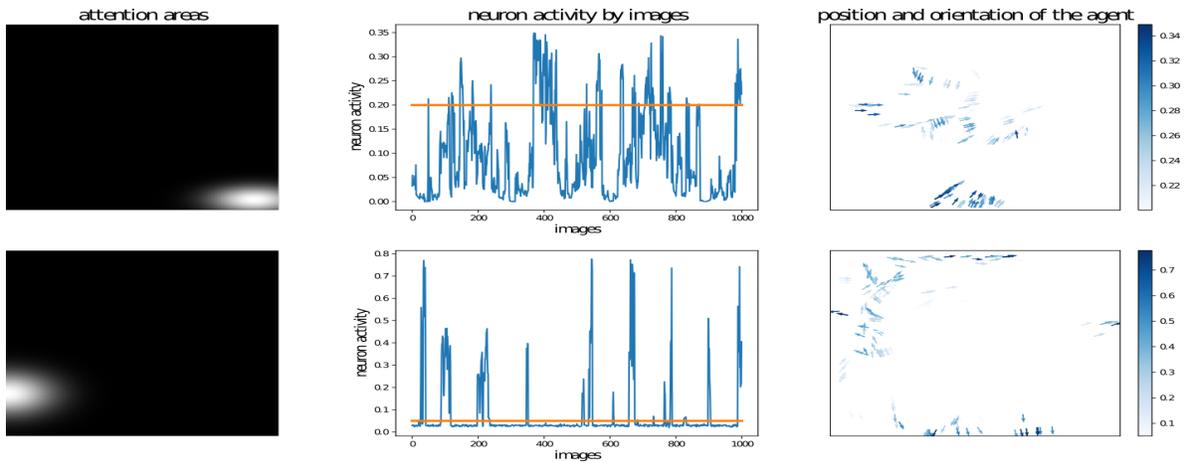}
    \caption{Study of the activity of two neurons, in both scenarios. Health gathering scenario on top and basic scenario at the bottom. The arrow on the last column represents the agent's position when the neuron activity is below the threshold (orange line), the color represents the amplitude of the activation.}
	\label{stm1}
\end{figure}

In figure \ref{stm1} we study two different neuron activations, one for each scenario, on 5000 images, and we print the position and orientation of the agent (black arrow) when the neuron is activated or inactivated under a certain threshold. In the basic scenario the neuron fires only when the agent has a wall at its left. Whereas, in the health gathering the neuron is activated when the agent is not facing a wall but rather when the floor is in the neuron attention area. Each aN gives a direct or indirect information about the environment, the monster can be identified by a neuron that gets activated on the wall; indeed, when this neuron will face the monster it will be deactivated indicating that there is something in front of the agent but it is not necessarily the monster. This is why a combination of different activations is needed to get a complete information.\\
All the neurons formed a pattern activation specific for each scenario, and inside this pattern there is small variation helping to differentiate states in the scenario. We show in the next section that this combination can provide enough information about the environment for Q-learning to predict relevant Q-values.

\section{Reinforcement Learning Application}
\label{sec5}
\begin{figure*}[!ht]
	
	\includegraphics[width=\textwidth,height=6cm]{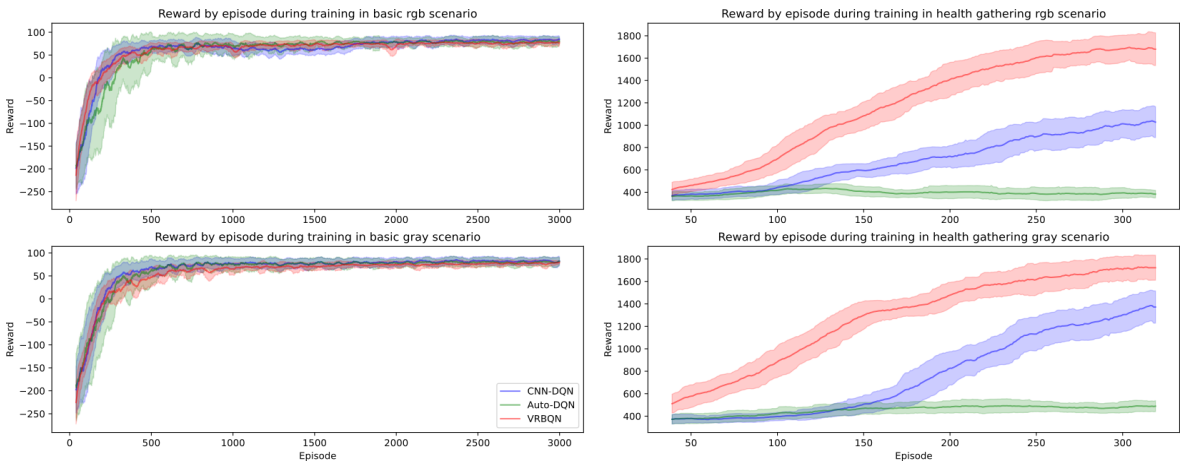}
	\caption{Comparison of 3 different approaches on basic (left) and health gathering (right) task on 20 different seeds. Rgb inputs for the top and gray for the bottom. Bold line represents the mean and transparent area the standard deviation. The plotted rewards for health gathering scenario is the number of alive step}
	\label{res}
	
\end{figure*}
In the following section we compare our algorithm on two different Vizdoom tasks, basic and health gathering scenarios, with a CNN-DQN, a convolutional network trained directly on pixels, and an autoencoder-DQN where the RL algorithm is applied on features extracted from a pre-trained CNN auto-encoder. Both CNN are similar to the one used in~\cite{Mnih2015} with one more convolutional layer as the inputs image are bigger, $120\times160$. Both scenarios have different characteristics, basic scenario has a sparse reward while health gathering reward is dense. Training is done with gray and RGB input state to check the robustness of tested networks to different input spaces. The dimension of the used images is $120\times160$. We stack two consecutive frames to form a state providing information about movement and we used a "skip frame" of 6 to train faster.\\
Agents are trained during 100k training steps. The RL part of CNN-DQN and autoencoder-DQN used target networks updated every 1000 iterations. An epsilon-greedy policy is used during training with epsilon starting to decay after 1000 training steps: it goes from 1 to 0.1 in 10 000 training step. A learning rate of 0.00025 and a batch size of 64 are used.\\
Our method, V-RBQN, does not require a target network as the features are sparse and the approximation of the Q-value is stable and can efficiently bootstrap. In addition, we did not use any exploration strategy and we used a learning rate of 0.01 and a batch size of 256.

\noindent For the basic scenario, the left curves in figure \ref{res} show that all the methods converge to the maximum reward. Both methods that trained the agent on extracted features have a larger variance during training than when training directly on images with convolution. This high variance is due to the imprecision of the features which can be too close for 2 different states that require 2 different actions to maximize the Q-value. For example when the monster is not perfectly in front of the weapon, the agent can be in a situation where the features are similar and so the Q-value for shoot, turn left and right are very close, the agent will oscillate between those three actions until it shoots the monster. In the health gathering scenario, which is more difficult to solve with Q-learning~\cite{Khan2020}, the auto-encoder method does not find an optimal solution and V-RBQN performs better that the CNN-DQN. Both state-of-the-art methods converge to a behavior where the agent can get stuck in walls, whereas our method alleviates this problem by having special local and sparse neurons activations.
Test results in table \ref{tab:T1} demonstrate the robustness of our method to different state spaces. Indeed, for the same number of training steps, our method gets similar results with different inputs size, i.e., corresponding to gray and rgb images, whereas CNN-DQN and autoencoder-DQN have a bigger variance and a smaller reward for rgb input when comparing with gray input for a given time-step. \\
\noindent Finally, we realized another test to highlight the sparsity of our network. We tested the learned weights only on the aNs for each scenario without the contribution of the iNs. Results in table \ref{tab:T1} show that the iNs are not required to solve a task. This allows faster forward passes as the quantity of neurons is divided by a factor of three when dealing with a specific scenario.

\begin{table}[t]
	\caption{Reward on 1000 episode of basic and health gathering scenario for 3 different algorithms. Score is calculated with mean$\pm$ standard deviation on 20 different seeds. V-RBQN (only aN) represents the reward when testing without inactive neurons.}
	\label{tab:T1}
	\begin{center}%
	\begin{tabu}{l X[c]X[c] X[c]X[c]}
		\toprule
		\multirow{2}{*}{Scenario} & \multicolumn{2}{c}{\texttt{basic}} & \multicolumn{2}{c}{\texttt{health gathering}}\\
		\cline{2-5}
		\texttt{} & \texttt{gray} & \texttt{rgb} & \texttt{gray} & \texttt{rgb}\\
		\midrule
		\midrule 
		\texttt{basic\-DQN} & $\mg{86\pm 6}$ & $ \mg{86\pm 6}$  & $1519\pm751$ & $1092\pm 716$  \\
		\texttt{Autoencoder-DQN} & $86\pm 7$ & $85\pm 7$ & $601\pm357$ & $438\pm 384$ \\
		\texttt{V-RBQN} & $85\pm8$ & $85\pm 9$ & $\mg{1809\pm543}$ & $\mg{1778\pm 573}$ \\
		\texttt{V-RBQN (only aN)} & $85\pm8$ & $84 \pm 9$ & $1799\pm 571$ & $1690\pm 635$ \\
		\bottomrule
	\end{tabu}
\end{center}
\end{table}

\section{Conclusion}
\label{sec6}
In this paper we have extended the Radial Basis Function Network to handle feature extraction in images. We put into light the sparsity and locality properties of our network by analyzing the extracted features. We then show the usefulness of these extracted features by using them as inputs of partially observable visual RL tasks.  
Despite its simplicity our Visual Radial Basis Q-Network (V-RBQN) gives promising results on two scenarios, i.e., the basic and the health gathering tasks, even with different input channel sizes. Thanks to the sparsity of the extracted features and their locality, the proposed approach outperforms the evaluated baseline while having less trainable parameters, without any exploration strategy nor target network. In addition, the inactive neurons can be omited without hindering the efficiency which allows to reduce the size of the network by a factor of three. One of the research direction will be to find generic RBF parameters, by training them on different types of images to be optimal regardless the input images.
Future work will also consider the extension of this approach to continuous action spaces and other RL algorithms.

\bibliographystyle{splncs04}
\bibliography{bib118}

\end{document}